**Title**

# Do LLMs produce texts with “human-like” lexical diversity?

**Authors and Affiliations**

Kelly Kendro[1*]
Jeffrey Maloney[2*]
Scott Jarvis[1]

[1] Northern Arizona University
[2] Brigham Young University – Hawaii

*These authors contributed equally to the manuscript.

**Corresponding address**

BYU–Hawaii #1954
55-220 Kulanui Street Bldg 5
Laie, Hawaii 96762-1293

Jeffrey.Maloney@byuh.edu

**Short title**

Lexical diversity in LLM-generated texts

**Acknowledgements**

The authors would like to thank two anonymous reviewers and attendees of the 2024 Annual Meeting of the Cognitive Science Society and the 2025 American Association for Applied Linguistics conference for their feedback and suggestions.

**Statements:**

- This work has not been published previously.
- This work is not under consideration for publication in another journal or elsewhere.
- All authors have seen, reviewed and contributed meaningfully to this work.
- The study received approval by the ethics committee of Northern Arizona University.
- We have not used any LLMs to generate the text of this manuscript.
- The data that support the findings of this study are available from the corresponding author upon reasonable request.

**Abstract**

The degree to which large language models (LLMs) produce writing that is truly human-like remains unclear despite the extensive empirical attention that this question has received. The present study addresses this question from the perspective of lexical diversity. Specifically, the study investigates patterns of lexical diversity in LLM-generated texts from four ChatGPT models (ChatGPT-3.5, ChatGPT-4, ChatGPT-o4 mini, and ChatGPT-4.5) in comparison with texts written by L1 and L2 English participants (*n* = 240) across four education levels. Six dimensions of lexical diversity were measured in each text: volume, abundance, variety-repetition, evenness, disparity, and dispersion. Results from one-way MANOVAs, one-way ANOVAs, and Support Vector Machines revealed that the ChatGPT-generated texts differed significantly from human-written texts for each variable, with ChatGPT-o4 mini and ChatGPT-4.5 differing the most. Within these two groups, ChatGPT-4.5 demonstrated higher levels of lexical diversity than older models despite producing fewer tokens. The human writers' lexical diversity did not differ across subgroups (i.e., education, language status). Altogether, the results indicate that ChatGPT models do not produce human-like texts in relation to lexical diversity, and the newer models produce less human-like texts than older models. We discuss the implications of these results for language pedagogy and related applications.

## Introduction

The introduction of public generative AI (GenAI) tools marked a new era, especially regarding text processing and production. With the growing popularity of these tools, there are questions about their abilities and limitations, especially regarding the extent to which their output resembles human writing. Major concerns include GenAI-facilitated plagiarism (e.g., Perkins, 2023) and the authenticity of using GenAI to generate teaching materials (e.g., Hu et al., 2025; Kasneci et al., 2023). There are a growing number of studies that have examined differences between human- and GenAI-generated texts. These include both subjective evaluations and identification of authorship (Casal & Kessler, 2023; Köbis & Mossink, 2021; Wen & Laporte, 2025) as well as systematic linguistic analysis to examine whether there are quantifiable differences between the language that GenAI versus humans produce (Berber Sardinha, 2024; Hamat, 2024; Herbold et al., 2023; Jiang & Hyland, 2024; Reviriego et al., 2024).

Multiple studies that have used human judgments have found that humans are not effective in distinguishing GenAI writing from human writing (Casal & Kessler, 2023; Köbis & Mossink, 2021; Wen & Laporte, 2025). However, given the heterogeneity of human writing, it remains unclear whether GenAI-generated texts measurably differ from human-written texts at different levels of education and English proficiency. The current study investigates the lexical diversity found in GenAI-generated texts and texts written by humans in response to an argumentative essay prompt to determine whether GenAI might produce language that is more representative of certain groups of human writers than of others.

## Literature Review

### *Lexical Diversity*

Lexical diversity is more than the opposite of word repetition (e.g., Jarvis 2013a, 2013b). Zipf (1935) was perhaps the first scholar to recognize this, describing lexical diversity (or what he called "rate of variegation") not only in relation to the overall amount of repetition in a text, but also in relation to word frequency distributions, intervals between repetitions of the same word, and the fact that lexical diversity is more closely associated with redundancy than with repetition. Zipf's observations highlighted the predictable and seemingly universal patterns of lexical diversity that can be found in large collections of texts. Later research emphasized that individual texts do not always show the same levels of lexical diversity depending on both text-internal (e.g., text length) and text-external (e.g., the vocabulary knowledge of the writer/speaker) variables (e.g., Carroll, 1938; Johnson, 1944; Yule, 1944). Indeed, in the decades since Zipf's seminal work, researchers have discovered that lexical diversity can reveal information about the person producing the text, such as their level of language proficiency (Díez-Ortega & Kyle, 2024; Woods et al., 2023), their language dominance in cases of bilingualism or multilingualism (e.g., Treffers-Daller and Korybski, 2015), their education level (González López & López-López, 2015; Martins, 2016), and whether they are a first (L1) or second (L2) language writer (e.g., González, 2017; Nasseri & Thompson, 2021).

Most research has operationalized lexical diversity using measures that produce estimates of the proportion of unique words versus repeated words in a text, and which do so in a way that compensates for the fact that the proportion of repeated words grows as the text becomes longer (e.g., Bestgen, 2024). Influenced by Zipf (1935), Jarvis (2013a) referred to these repetition-based indices as measures of variability and argued that they represent only one of many dimensions of lexical diversity that collectively affect how human judges perceive this phenomenon. To avoid terminological confusion, Akbary and Jarvis (2023) replaced the term *variability* with *variety*– *repetition* and listed the other six dimensions of lexical diversity as follows: volume (text length), abundance (number of unique words), evenness (the text's word frequency distribution), dispersion (intervals between repetitions of the same word), disparity (degree of difference between the words in the text), and specialness (the presence of particular words that markedly enhance the text's perceived lexical diversity). These dimensions not only affect perceptions of lexical diversity (Jarvis, 2012, 2013a, 2017; Vanhove et al., 2019) but also reflect the characteristics of lexical diversity observed by Zipf (1935) as well as the multidimensional models and measures of diversity adopted by the field of ecology (e.g., Chao & Jost, 2012). The present study makes use of six of these seven dimensions of lexical diversity, as described in our Method section.

***Human judges' ability to differentiate between human- and GenAI-produced writing***

Researchers have investigated whether humans can successfully discern whether a given text was written by a human or a machine. Köbis and Mossink (2021) asked human judges to try to distinguish between poetry written by humans versus an early GPT model (GPT-2); the participants were not generally able to do so. Casal and Kessler (2023) examined whether applied linguists could reliably identify whether abstracts for four different studies were human- or GenAI-written. Again, even highly-trained applied linguists were not successful in discerning authorship. Yeadon et al. (2024) asked participants to judge whether short-essay responses to physics questions were GenAI- or human-written. Here, too, participants were unsuccessful in correctly identifying GenAI-generated texts. More recently, Wen and Laporte (2025) asked human raters to try to distinguish between experiential narratives written by humans versus two GenAI models (ChatGPT-3.5 and ChatGPT-4.0). Though ChatGPT-3.5 and ChatGPT-4.0 produced texts that were respectively less and more lexically diverse than human-written texts, the participants were still unable to accurately distinguish between the GenAI- and human-written narratives.

***Linguistic analyses of GenAI-produced language***

GenAI uses large language model (LLM) technology, computational neural networks typically based on the transformer architecture. LLMs are trained on vast amounts of language data for the purpose of learning patterns of language use and creating models that allow them to predict relationships among words, phrases, and larger units of language (Raiaan et al., 2024). As discussed above, they have been found to be exceedingly successful in achieving human-like language interpretation and production. ChatGPT, OpenAI's proprietary closed-source model with a chatbot interface, is perhaps the most ubiquitous of these tools, situated at the intersection of generative models, LLMs, artificial neural networks, and chatbots (Guest et al., 2025: 5-6). The present paper follows up on the finding that, though human judges often fail to distinguish between human- and GenAI-generated language, detailed linguistic analysis is more successful (e.g., Mizumoto et al., 2024).

Several studies have conducted detailed linguistic analyses comparing human and GenAI writing across registers. For example, Berber Sardinha (2024) used Biber's (1995) five dimensions of register variation to analyze a corpus of texts generated by ChatGPT-3.5 across a variety of genres. The length-matched results were then compared to human produced texts from a variety of existing corpora, and a discriminant function analysis found the multidimensional analysis to have high predictive power for determining whether a text was written by humans or an LLM. Similarly, Akinwande et al. (2024) found NLP and statistical analyses (i.e., sentiment analysis, theme analysis, vocabulary diversity) to accurately identify human versus GenAI authorship. Other researchers have analyzed the lexical diversity dimension of variety–repetition, finding it to reliably discriminate between human-written and GenAI-generated writing across registers such as academic essays (Herbold et al., 2023; Jiang & Hyland, 2025), TOEFL prompt responses (Reviriego et al., 2024), academic articles (Tudino & Qin, 2024), and poetry (Hamat, 2024). Across these studies, GenAI-written texts from ChatGPT models (i.e., ChatGPT-3, ChatGPT-3.5, and/or ChatGPT-4.0) varied in whether their lexical diversity was higher or lower than human-written texts, with older models often having lower comparative diversity and newer models having higher comparative diversity (e.g., Herbold et al., 2023; Reviriego et al., 2024), yet still producing consistently *different* levels of lexical diversity than human writing.

Other studies have examined the impact of incorporating GenAI writing tools into the writing process. Baseline use of GenAI has been shown to increase lexical diversity and result in longer texts (Kim & Chon, 2025), though Padmakumar and He (2024) found that different interfaces (e.g., base GPT3 versus a feedback-tuned model built on the GPT3 framework) may lead to different levels of lexical diversity. These differences appear to be widespread as researchers integrate GenAI into their writing. Bao et al. (2025) evaluated the linguistic changes in research abstracts before and after the availability of GenAI. Pre-GenAI, there was a larger gap between abstracts authored by L1 and L2 English writers. Abstracts authored in the GenAI era showed more similar lexical complexity, clarity, and cohesion between these groups, though readability declined due to increased use of jargon. Similarly, Lin et al. (2025) found that L2 English abstracts demonstrated a marked increase in lexical diversity (i.e., variety–repetition). Kwok et al. (2025) found further that translations produced by students in Hong Kong were more lexically sophisticated and diverse when created in conjunction with ChatGPT. These findings echo other studies that have found that utilizing ChatGPT to assist in writing produced longer and more lexically diverse texts.

### *The Current Study*

This exploratory study contributes to the ongoing systematic analysis and comparison of human and GenAI writing by examining whether ChatGPT models produce texts that are truly “human-like” from the perspective of lexical diversity when compared to L1 and L2 human writing across several education levels for one common use case: argumentative essay writing. We ask the following research questions:

- RQ1: Does the lexical diversity of ChatGPT-generated texts differ from that of human-generated texts?
- RQ2: Do different ChatGPT models produce differing levels of lexical diversity?
- RQ3: Do human writers produce differing levels of lexical diversity depending on their levels of education and/or their L1/L2 status?
- RQ4: Are texts generated by ChatGPT more similar to the writing of certain groups of human writers than to others, in relation to their lexical diversity profiles?

## Method

### *Materials and Procedure*

To answer our research questions, we presented human participants with a prompt used in the 2006 and 2007 TOEFL examination, “Do you agree or disagree with the following statement? It is better to have broad knowledge of many subjects than to specialize in one specific subject. Please write 250 or more words and use specific reasons and examples to support your answer.” We then presented this prompt to four versions of ChatGPT (ChatGPT-3.5, ChatGPT-4.0, ChatGPT-o4 mini, and ChatGPT-4.5), additionally directing the LLM to “provide an essay response in prose.” This involved zero-shot prompting (i.e., no examples provided in the prompt) and no manipulation of the ChatGPT models’ temperature (i.e., its creativity parameter). We prompted each model thirty times, collecting a total of 120 responses from all four models.

The 240 human participants (described in the following section) not only responded to the prompt but also completed a short demographic survey and self-rated their English writing ability, writing complexity, writing fluency, and difficulty producing written texts. Participants who reported that English was not their dominant language completed a C-test (Ishihara et al., 2003) as a measure of English proficiency and self-rated their L1 writing ability, complexity, fluency, and the difficulty they experience producing written texts.

***Participants***

Our human participants were recruited from Prolific (Prolific.com), an online subject pool. Prolific's payment structure evaluates participants' mean and median completion times in relation to the flat rate payment, and data collection is paused if compensation falls below a minimum threshold, requiring additional compensation for future and past participants to resume data collection. As such, there is no incentive for participants to complete tasks quickly, as longer completion times will result in higher payment.

As Table 1 shows, we collected data from a total of 240 people, evenly split across L1 and L2 English users per their responses to Prolific screening questions. Within those two groups of 120, we recruited 30 participants at each of our four target education levels: high school, bachelor's, master's, and doctoral degrees. These are the highest levels of education that they had completed prior to participating in the study. We had 91 male, 143 female, and 6 nonbinary participants. Their ages ranged from 18 to 79 ($M$ = 38.99, $SD$ = 12.73).

Table 1. Human participants by education level and L1/L2 status

| Education completed | L1 English | L2 English | TOTAL |
| --- | --- | --- | --- |
| High School | 30 | 30 | 60 |
| Bachelor's Degree | 30 | 30 | 60 |
| Master's Degree | 30 | 30 | 60 |
| Doctorate | 30 | 30 | 60 |
| TOTAL | 120 | 120 | 240 |

### ***Analysis***

All 360 texts (i.e., 120 essays generated by ChatGPT models and 240 texts written by human participants) were lemmatized using TreeTagger (Schmid, 1994) and were then submitted to lexical diversity analysis using Python scripts created by the authors. The Python scripts produced the following six lexical diversity indices for each text (cf. Akbary & Jarvis, 2023):

- Volume: number of occurrences of individual word tokens in the text,
- Abundance: number of word types in the text, where types were operationalized as word lemmas (i.e., all inflections of the same word were counted as tokens of the same type),
- Variety–repetition: measured with the MATTR index (Covington & McFall, 2010), which outputs the mean TTR of all consecutive sequences of 50 words in a text,
- Evenness: measured with an evenness index derived from Shannon’s diversity index (Jost, 2010); the evenness measure reflects how evenly the tokens in a text are distributed across types,
- Disparity: relies on the WordNet sense index and measures the mean number of words in a text that share a meaning sense (Jarvis, 2013),

- Dispersion: measures the relative number of times that tokens of the same type are repeated within 20 words of one another; repetitions that occur in such close proximity reflect the opposite of dispersion, so the values for this measure are on an inverse scale (Akbary & Jarvis, 2023).

These measures represent six of the seven properties that Jarvis (2013a; 2013b; 2017) has argued as constituting the full construct of lexical diversity. They collectively predict human perceptions of lexical diversity (Jarvis, 2013a, 2017; Vanhove et al., 2019) and ultimately reflect the lexical dimensions along which discourse is structured to optimally hold the reader's (or listener's) interest and attention while providing enough repetition and redundancy at the appropriate intervals to keep the discourse coherent and comprehensible (Zipf, 1935). Dispersion is specifically a measure of the intervals between repetitions of the same word (lemma), whereas disparity gauges the degree of semantic redundancy in the text when the writer (or speaker) reinforces the lexical cohesion of the text through synonyms rather than through repetitions of the same word (see, e.g., Halliday & Hasan, 1976). Evenness is useful for identifying texts where the repetition of some words is out of balance with the text's overall pattern of word repetition, for example, in ways that violate the Zipfian distribution (Zipf, 1935). Variety-repetition is the property of lexical diversity that is measured in most of the existing research on lexical diversity. It specifically measures the relationship between volume (tokens) and abundance (types) and reflects the proportion of unique words to repeated words in a text. Variety-repetition, volume, and abundance have each been found to correlate highly with perceptions of lexical diversity (e.g., Vanhove et al., 2019) and assessments of writing quality and language proficiency (e.g., Treffers-Daller, 2013; Yu, , 2020; Zenker & Kyle, 2021), but the combination of the whole set of lexical diversity properties provides even stronger predictions of writing quality and perceptions of lexical diversity (e.g., Jarvis, 2017; Vanhove et al., 2019). For these reasons, we are including all six of these lexical diversity measures in the present study, as the literature suggests that all six are relevant to lexical diversity, lexical cohesion, coherence, and writing quality.

The statistical tests used to address the research questions included one-way MANOVAs, permutations-based MANOVAs, one-way ANOVAs, Kruskall–Wallis tests, relevant post-hoc tests, and Support Vector Machines. All statistical analyses were conducted in R version 4.5.1 (R Core Team, 2025).

**Results**

To address RQ1, a one-way MANOVA was conducted to examine whether there were overall differences in lexical diversity between texts generated by ChatGPT vs. those written by humans. The six dependent variables were volume, abundance, MATTR, evenness, disparity, and dispersion. Appendix Table 1 shows the means, confidence intervals, standard deviations, and minimum and maximum values for the combined ChatGPT- vs. human-generated texts in relation to all six lexical diversity measures. The MANOVA results revealed a statistically significant multivariate effect of writer type (i.e., human vs. ChatGPT), *Wilks' Λ* = .181, $F(6, 353) = 267.06$, $p < .001$. Additional multivariate test statistics corroborated the significance of writer type, including Pillai's Trace = 0.819, Hotelling-Lawley Trace = 4.539, and Roy's Largest Root = 4.539 (see Table 2). The effect size for the overall multivariate effect was large, as indicated by Wilks' Lambda partial $\eta^2 = 0.819$, demonstrating that approximately 82% of the variance in the combined lexical diversity measures was explained by writer type. Due to a significant violation of normality (Shapiro-Wilks $W = .937$, $p < .001$), we ran a permutations-based MANOVA (adonis2) on the data using 999 permutations with the Euclidean method. The results of the permutations-based MANOVA (Sum of Squares = 4,094,593, $R^2 = .72$, $F = 923$, $p < .001$) confirmed the significant effects observed in the parametric MANOVA.

Table 2. MANOVA model fit statistics: Human vs. ChatGPT

| Fit measure | *df* | Approx. *F* | Coefficient | *p*-value | Residuals |
|---|---|---|---|---|---|
| Wilks | 6/353 | 267.07 | 0.181 | < .001 | 358 |
| Pillai | 6/353 | 267.06 | 0.819 | < .001 | 358 |
| Hotelling–Lawley | 6/353 | 267.06 | 4.539 | < .001 | 358 |
| Roy | 6/353 | 267.06 | 4.539 | < .001 | 358 |

Post-hoc one-way ANOVAs showed that the differences between ChatGPT- and human-generated texts were significant ($p < .001$) with respect to each of the six lexical diversity measures. Partial $\eta^2$ squared values ranged from medium to large effects, with confidence intervals supporting the robustness of these effects (see Table 3). Given the significant violation of normality in our data, we also ran the Kruskal-Wallis nonparametric test for each dependent variable. This yielded results consistent with our parametric ANOVA, as shown in Table 3. According to Cohen's (1988) interpretation guidelines, the effect size for disparity was medium, whereas all others were large. Appendix figures A1–A6 provide visual representations of these differences as violin plots with embedded boxplot elements, illustrating substantially higher statistical distributions for the ChatGPT group than for the human group across all lexical diversity measures except dispersion, where the ChatGPT distribution is substantially lower. As dispersion's scale is inverted, lower values represent higher levels of lexical diversity; collectively, the results of this analysis show that ChatGPT-generated texts have significantly and substantially higher levels of lexical diversity than human-generated texts for all six measures.

Table 3. Post-hoc ANOVA results comparing ChatGPT- and human-generated texts on each dependent variable

| LD measure | $F(1, 358)$ | *p*-value | partial $\eta^2$ | 95% CI | Kruskall-Wallis | *p*-value (nonparametric) |
|---|---|---|---|---|---|---|
| Volume | 887.129 | < .001 | .712 | [.68, 1.00] | 224.41 | < .001 |
| Abundance | 1039.875 | < .001 | .744 | [.71, 1.00] | 236.81 | < .001 |
| MATTR | 476.585 | < .001 | .571 | [.52, 1.00] | 207.22 | < .001 |
| Evenness | 288.603 | < .001 | .446 | [.39, 1.00] | 165.67 | < .001 |
| Disparity | 51.010 | < .001 | .125 | [.08, 1.00] | 47.466 | < .001 |
| Dispersion | 568.855 | < .001 | .614 | [.57, 1.00] | 222.67 | < .001 |

To address RQ2, we limited our analysis to the 120 ChatGPT-produced texts, using a one-way MANOVA to determine whether there were overall differences in lexical diversity between texts generated by four ChatGPT models. The MANOVA results revealed a statistically significant multivariate effect of ChatGPT version, *Wilks' Λ* = .020, $F(3, 116) = 51.60$, $p < .001$, partial $\eta^2 = .98$. This result was confirmed with a permutations-based MANOVA (adonis2) using 999 permutations, Sum of Squares = 839,875, $R^2 = .69$, $F = 84.77$, $p < .001$.

Post-hoc one-way ANOVAs with Tukey pairwise comparisons and Bonferroni corrections showed that the significant differences between ChatGPT models were as shown in Table 4) with respect to each measure of lexical diversity.

Table 4. Post-hoc homogeneous subsets of ChatGPT-generated texts by lexical diversity variable

| Variable | Comparative groupings of ChatGPT-generated texts |
| --- | --- |
| Volume | ChatGPT-4.5 < ChatGPT-3.5 = ChatGPT-o4 mini < ChatGPT-4.0 |
| Abundance | ChatGPT-4.5 = ChatGPT-3.5 < ChatGPT-4.0 < ChatGPT-o4 mini |
| MATTR | ChatGPT-3.5 = ChatGPT-4.0 < ChatGPT-4.5 < ChatGPT-o4 mini |
| Evenness | ChatGPT-3.5 = ChatGPT-4.0 < ChatGPT-4.5 = ChatGPT-o4 mini |
| Disparity | ChatGPT-4.5 = ChatGPT-3.5 = ChatGPT-o4 mini = ChatGPT-4.0 |
| Dispersion | ChatGPT-o4 mini < ChatGPT-4.5 < ChatGPT-4.0 = ChatGPT-3.5 |

These results can be summarized as follows. ChatGPT-4.0 produced the longest texts and ChatGPT-4.5 produced the shortest. Differences in text length also affected their abundance levels. There were no significant differences between ChatGPT-3.5 and ChatGPT-4.0 for any of the remaining four measures, which do not vary by text length (cf. Bestgen, 2024). For three of these remaining measures (MATTR, evenness, and dispersion), the two newest versions of ChatGPT (ChatGPT-o4 mini and ChatGPT-4.5) showed the highest levels of lexical diversity, with ChatGPT-o4 mini significantly higher than ChatGPT-4.5 for both MATTR and (inverted scale) dispersion.

For RQ3, we limited our analysis to the 240 texts produced by human writers, divided by education level and L1/L2 status into eight groups of 30 writers per group, as shown earlier in Table 1. A one-way MANOVA did not find a statistically significant multivariate effect of group, *Wilks' Λ* = .824, $F(7, 232) = 1.070$, $p = .353$, partial $\eta^2 = .18$. The lack of a statistical difference across groups and measures was confirmed with a 999-permutation MANOVA (adonis2), Sum of Squares = 13,518, $R^2 = .37$, $F = 1.27$, $p = .23$.

Additionally, we investigated whether writing ability or age significantly predicted the number of types or tokens produced. Linear regression models with number of types and tokens as outcome variables and age and self-reported perceived ability to write well, ability to write fluently, writing complexity/sophistication, and difficulty writing found that none of these participant variables significantly predicted number of tokens ($F(5,230) = 1.648$, $p = .148$, $R^2 = .014$), and that only writing complexity/sophistication marginally significantly predicted number of types ($F(5,230) = 2.683$, $p = .022$, $R^2 = .035$).[1]

Our final research question (RQ4) can be paraphrased as follows: Does at least one model produce lexical diversity values similar to at least one of the groups of human writers? Table 5 shows means and standard deviations for each group's lexical diversity scores. We used a series of one-way ANOVAs combined with Tukey post-hoc tests and Bonferroni corrections to identify specific pairs of human groups and ChatGPT models that did not differ significantly with respect to one or more of our lexical diversity measures. The Bonferroni-corrected *p*-values showed that each human group differed significantly from every ChatGPT model with respect to all measures of lexical diversity except disparity. For the disparity measure, no group of human writers differed significantly from ChatGPT-3.5 or ChatGPT-4.5. Also, the L1 writers who had completed a bachelor's degree did not differ significantly from ChatGPT-o4 mini with respect to disparity. As a reminder, disparity in the present study is operationalized as the proportion of types in a text that share a meaning sense, so these results show that ChatGPT-3.5 and ChatGPT-4.5 produced relatively human-like patterns with respect to the relative proportions of synonyms they produced per text.

---

[1] We calculated the VIF to evaluate whether multicollinearity was an issue in these models, finding that there were no tolerances lower than .33 and no VIFs above 2.99, thus indicating that the predictors were appropriate to use together.

Table 5. Mean (standard deviation) for each group and ChatGPT models for all lexical diversity measures

| Group | Volume | Abundance | MATTR | Evenness | Disparity | Dispersion |
|---|---|---|---|---|---|---|
| L1 HS | 274.43 (33.73) | 129.00 (21.49) | 38.38 (2.08) | 0.97 (0.01) | 1.03 (0.01) | 16.82 (5.04) |
| L2 HS | 269.83 (28.48) | 126.77 (20.82) | 38.31 (2.09) | 0.97 (0.01) | 1.03 (0.01) | 16.37 (4.66) |
| L1 BA | 279.57 (49.24) | 128.07 (19.15) | 38.43 (1.69) | 0.97 (0.00) | 1.03 (0.01) | 16.58 (3.17) |
| L2 BA | 267.83 (16.41) | 128.33 (14.19) | 38.72 (1.57) | 0.97 (0.01) | 1.03 (0.01) | 16.07 (4.35) |
| L1 MA | 269.27 (25.84) | 132.97 (18.84) | 38.57 (2.06) | 0.97 (0.01) | 1.03 (0.01) | 16.20 (4.35) |
| L2 MA | 265.77 (22.31) | 132.13 (17.42) | 38.45 (2.08) | 0.97 (0.01) | 1.03 (0.01) | 16.41 (4.35) |
| L1 PhD | 287.07 (51.59) | 138.23 (29.55) | 38.33 (1.99) | 0.97 (0.01) | 1.03 (0.01) | 16.41 (3.68) |
| L2 PhD | 270.30 (16.78) | 132.27 (14.55) | 38.73 (1.63) | 0.97 (0.01) | 1.03 (0.01) | 16.25 (3.65) |
| GPT-3.5 | 468.30 (35.24) | 213.30 (15.39) | 41.28 (0.57) | 0.98 (0.00) | 1.04 (0.01) | 8.65 (1.16) |
| GPT-4.0 | 542.60 (37.84) | 263.33 (19.06) | 41.52 (0.64) | 0.98 (0.00) | 1.04 (0.01) | 8.23 (1.21) |
| GPT-4.5 | 349.93 (34.48) | 207.17 (13.36) | 44.18 (0.81) | 0.99 (0.00) | 1.04 (0.01) | 5.81 (1.15) |
| GPT-o4 mini | 501.30 (83.58) | 313.47 (39.71) | 44.97 (0.63) | 0.99 (0.00) | 1.04 (0.01) | 4.81 (1.11) |

To complement our group-level analyses (i.e., MANOVA and ANOVA) with an approach that assesses whether individual texts can be classified as human- or ChatGPT-generated based on their lexical diversity profiles, we also addressed RQ1 using a Support Vector Machine (SVM) analysis. SVM is an optimization-based margin maximization classifier that does not assume normality of the data, equal variances across groups, independence of features, or any specific functional form for the distribution of predictors (Suthaharan, 2016). Like other classifiers such as linear discriminant analysis and logistic regression, it uses independent variables (measures of lexical diversity, in this case) to predict group (e.g., human or ChatGPT) to which a particular sample (i.e., text) belongs. Because we were not interested in the classifier's ability to determine a text's group membership based on how long the text is, we removed from this stage of our analysis the two lexical diversity measures that reflect text length: volume and abundance. Our SVM model was thus trained to classify texts as generated by ChatGPT or by humans using only the following four lexical diversity predictors: MATTR, evenness, disparity, and dispersion. Our SVM analysis relied on the caret (Kuhn et al., 2020) and e1071 (Meyer et al., 2024) packages in R using a radial basis function (RBF) kernel, which is a non-linear SVM kernel capable of capturing more complex patterns than a linear SVM.

To ensure robust evaluation while minimizing overfitting, the dataset was first split using stratified sampling into training/validation and test sets, with 80% allocated to training/validation and 20% reserved for testing. Subsequently, the training/validation subset was further divided into 80% for training and 20% for validation, resulting in an approximate overall split of 64% training, 16% validation, and 20% test. Model training included 5-fold cross-validation within the training set for hyperparameter tuning, optimizing the model based on the area under the receiver operating characteristic curve (ROC-AUC). Model tuning, validation, and final evaluation were conducted on separate data partitions in order enhance the reliability and generalizability of the results.

Performance metrics were evaluated across 10 different random seeds to assess model stability. The average F1 score was 0.969 (SD = 0.015), and the average ROC-AUC was 0.989 (SD = 0.008), with the lower bounds of the confidence intervals consistently exceeding 0.94, indicating high levels of discrimination between ChatGPT-generated and human-written texts. Table 6 presents performance metrics by seed, including F1 scores, ROC-AUC values, and their corresponding 95% confidence intervals, along with summary statistics (mean and standard deviation).

Table 6. SVM performance metrics during training and validation: Human vs. ChatGPT

| Seed | F1 | AUC | LowerCI | UpperCI |
|---|---|---|---|---|
| 2479 | 0.958 | 0.988 | 0.970 | 1.000 |
| 4155 | 0.957 | 0.990 | 0.973 | 1.000 |
| 7016 | 0.957 | 0.974 | 0.943 | 1.000 |
| 1093 | 0.989 | 0.992 | 0.976 | 1.000 |
| 153 | 0.979 | 0.992 | 0.978 | 1.000 |
| 4556 | 0.945 | 0.977 | 0.950 | 1.000 |
| 3779 | 0.968 | 0.990 | 0.972 | 1.000 |
| 7238 | 0.989 | 0.999 | 0.997 | 1.000 |
| 3776 | 0.979 | 0.993 | 0.979 | 1.000 |
| 8270 | 0.968 | 0.992 | 0.980 | 1.000 |
| **Mean** | 0.969 | 0.989 | – | – |
| **SD** | 0.015 | 0.008 | – | – |

The final SVM model demonstrated strong classification performance on the held-out test set of 72 texts (i.e., 20% of the full dataset of 360 texts). Key metrics of model performance include an F1 score of 0.968, reflecting a balanced trade-off between precision (the proportion of predicted positives that are true positives) and recall/sensitivity (the proportion of actual positives correctly detected). Precision was perfect at 1.000, indicating that all samples predicted as the positive class (Human) were indeed true positives. Sensitivity was high at 0.938, showing that the model correctly identified most of the actual positives. Specificity was perfect at 1.000, indicating the model correctly identified all actual negatives (ChatGPT). The confusion matrix in Table 7 shows that the final SVM model achieved an overall accuracy of 95.8% (95% CI[91.2%, 100%]), correctly identifying all 45 test-set texts written by humans, and correctly identifying 24 (88.9%) of the 27 texts generated by ChatGPT.

Table 7. SVM final model confusion matrix: Human vs. ChatGPT

| | | Predicted | | |
|---|---|---|---|---|
| | | ChatGPT | Human | % Accuracy |
| Observed | ChatGPT | 24 | 3 | 88.9% |
| | Human | 0 | 45 | 100.0% |
| Overall Prediction Accuracy | | | | 95.8% |

The overall contributions of the lexical diversity features to the SVM model's predictions were assessed using SHAP values, which quantify the average impact of each feature on the model's output across the test set. Dispersion emerged as the most influential predictor with a mean absolute SHAP value of 0.279, indicating that it had the strongest effect in distinguishing human-written from ChatGPT-generated texts. MATTR was the second most important feature (mean absolute SHAP = 0.135), followed by disparity (0.055) and evenness (0.036).

We ran a second SVM analysis to address our remaining research questions (RQ2–RQ4) from the perspective of text classification. The second SVM, like the first, included all 360 texts in our database and used the same four measures of lexical diversity (dispersion, MATTR, disparity, and evenness) as predictors. However, this time the group variable divided the database into 12 groups of texts depending on whether each text was written by a human vs. ChatGPT, which version of ChatGPT (if any) produced the text, whether the human writer (in relevant cases) was an L1 vs. L2 speaker of English, and what the human writer's level of education was. Each group consisted of 30 texts (see Table 8).

Table 8. Levels of the group variable used in the second SVM

| Group | Human vs. ChatGPT | L1/L2/model status | Education level | *n* |
|---|---|---|---|---|
| 1 | Human | L1 English | high school | 30 |
| 2 | Human | L2 English | high school | 30 |
| 3 | Human | L1 English | bachelor's | 30 |
| 4 | Human | L2 English | bachelor's | 30 |
| 5 | Human | L1 English | master's | 30 |
| 6 | Human | L2 English | master's | 30 |
| 7 | Human | L1 English | doctorate | 30 |
| 8 | Human | L2 English | doctorate | 30 |
| 9 | ChatGPT | ChatGPT-3.5 | – | 30 |
| 10 | ChatGPT | ChatGPT-4.0 | – | 30 |
| 11 | ChatGPT | ChatGPT-4.5 | – | 30 |
| 12 | ChatGPT | ChatGPT-o4 mini | – | 30 |

As before, the SVM proceeded through stratified sampling, with 80% of the texts allocated to training/validation and 20% reserved for testing. The training/validation subset was further divided into 80% for training and 20% for validation, with a final approximate overall split of 64% training, 16% validation, and 20% test. Like the first, the second SVM analysis involved 5-fold cross-validation, with model tuning, validation, and final evaluation conducted on separate data partitions. Performance metrics derived from 10 random seeds showed a macro-averaged F1 score of 0.249 (*SD* = 0.032) and an average ROC-AUC of 0.661 (*SD* < 0.001).

The final model of the second SVM analysis showed F1 scores ranging from 0.111 (Group 4) to 0.471 (Group 12). However, F1 values could not be calculated for seven of the 12 groups due to the fact that there were no positive predictions for these seven groups. Similarly, as shown in the confusion matrix in Table 9, the final SVM model achieved 0.0% accuracy in relation to eight of the 12 groups, with an overall prediction accuracy of 16.7%, 95% CI[0.089, 0.273]. Although this is significantly above the 8.33% level of chance ($p = 0.015$), the SVM model was not generally able to distinguish texts by the backgrounds of the human writers or by the versions of ChatGPT used. We note, however, that the two groups with the highest prediction accuracy (i.e., 50%) both represented high school graduates (L1 and L2 English). Also worthy of note is the fact that most misclassifications were within human groups or within ChatGPT versions. The only exceptions to this relate to Group 8 (human L2 English writers with doctoral-level education), Group 9 (ChatGPT 3.5), and group 10 (ChatGPT 4.0). Six of the 12 texts written by members of Group 8 were misclassified as having been produced by ChatGPT, four of the six texts produced by ChatGPT 3.5 were misclassified as having been written by humans, and three of the 12 texts produced by ChatGPT 4.0 were also misclassified as having been written by humans.

Table 9. SVM final model confusion matrix: All 12 groups

| | | Predicted Group | | | | | | | | | | | | |
|---|---|---|---|---|---|---|---|---|---|---|---|---|---|---|
| | | *1* | *2* | *3* | *4* | *5* | *6* | *7* | *8* | *9* | *10* | *11* | *12* | Accuracy |
| Observed | *1* | **3** | 1 | 0 | 1 | 0 | 0 | 0 | 0 | 0 | 0 | 0 | 1 | 50.0% |
| Group | *2* | 1 | **3** | 0 | 1 | 0 | 0 | 0 | 1 | 0 | 0 | 0 | 0 | 50.0% |

| | | | | | | | | | | | | | |
|---|---|---|---|---|---|---|---|---|---|---|---|---|---|
| *3* | 0 | 3 | **0** | 2 | 0 | 0 | 0 | 1 | 0 | 0 | 0 | 0 | 0.0% |
| *4* | 1 | 3 | 0 | **1** | 0 | 0 | 0 | 1 | 0 | 0 | 0 | 0 | 16.7% |
| *5* | 0 | 5 | 0 | 1 | **0** | 0 | 0 | 0 | 0 | 0 | 0 | 0 | 0.0% |
| *6* | 0 | 3 | 0 | 3 | 0 | **0** | 0 | 0 | 0 | 0 | 0 | 0 | 0.0% |
| *7* | 3 | 1 | 0 | 2 | 0 | 0 | **0** | 0 | 0 | 0 | 0 | 0 | 0.0% |
| *8* | 1 | 3 | 0 | 1 | 0 | 0 | 0 | **1** | 4 | 0 | 2 | 0 | 8.3% |
| *9* | 0 | 0 | 0 | 0 | 0 | 0 | 0 | 4 | **0** | 0 | 2 | 0 | 0.0% |
| *10* | 0 | 0 | 0 | 0 | 0 | 0 | 0 | 3 | 1 | **0** | 2 | 6 | 0.0% |
| *11* | 0 | 0 | 0 | 0 | 0 | 0 | 0 | 0 | 0 | 0 | **0** | 4 | 0.0% |
| *12* | 2 | 0 | 0 | 0 | 0 | 0 | 0 | 0 | 0 | 0 | 0 | **0** | 0.0% |
| | | | | | | | | | Overall Prediction Accuracy | | | | 16.7% |

**Discussion**

The current study aimed to compare the lexical diversity of argumentative essay texts generated by four ChatGPT models to texts written by human participants from different educational and language backgrounds. We collected data from 240 human subjects across eight groups, balanced for language status (L1/L2 English) and educational level (high school, bachelor's, master's, and doctorate), and analyzed six dimensions of lexical diversity for each text. Our research questions asked whether the lexical diversity of ChatGPT and human writing broadly differed, whether lexical diversity varied within these two groups (i.e., between ChatGPT models and between human writer groups), and whether any of the ChatGPT models demonstrated lexical diversity profiles similar to humans from a particular group for this task.

***Human versus ChatGPT writing***

Our first research question asked whether humans and ChatGPT models produced different levels of lexical diversity in their responses. The MANOVA and ANOVA tests showed clear distinctions between ChatGPT and human writing for each of the six lexical diversity dimensions, indicating that ChatGPT produces starkly different levels of lexical diversity than humans regardless of model. The results of the Support Vector Machine analysis revealed that dispersion and variety-repetition contributed the most to the statistical model. The differences between texts produced by humans vs. ChatGPT were thus confirmed at both the group level and at the level of individual texts, indicating that ChatGPT models produced different patterns of lexical diversity. While our results confirm previous findings regarding differences in lexical diversity between human- and LLM-generated texts, our results consistently demonstrate that ChatGPT-generated writing had higher lexical diversity than the human-written texts. This contrasts with previous work that has found LLM-generated texts to have lower (Hamat, 2024; Tudino & Qin, 2024) or more variable (e.g., some higher, some lower; Herbold et al., 2023; Reviriego et al., 2024) lexical diversity than human-written texts.

***Lexical diversity within models***

Our second research question asked whether individual ChatGPT models would produce different levels of lexical diversity. To answer this question, we investigated four ChatGPT models and found that there was some variation in the models. ChatGPT-o4 had the highest and most distinctive levels of lexical diversity, while ChatGPT-3.5 and ChatGPT-4.0 performed similarly (cf. Reviriego et al., 2024; Wen & Laporte, 2025). Regarding individual lexical diversity dimensions, we found that disparity did not differ across models, while the other five dimensions (variety, abundance, variety–repetition, evenness, and dispersion) did. Despite these between-model differences, the LLMs patterned similarly to each other when compared to human writing.

***Lexical diversity across populations***

In RQ3, we investigated lexical diversity within the eight human participant groups. In addition to investigating differences both in language status (i.e. L1 versus L2 English) and education level (i.e., high school versus bachelor's versus master's versus doctorate), we used a MANOVA to analyze lexical diversity scores between each of the language status X education level groups. Though we had expected to see some variation, the participant groups did not differ from one another in terms of their lexical diversity scores, a finding that held for all six dimensions analyzed. These results were unexpected, yet they are somewhat consistent with previous literature. Our finding that there were no meaningful differences in lexical diversity scores between L1 and L2 writers confirms the findings of some past research (Kyle et al., 2021; Nasseri & Thompson, 2021) but is at odds with other studies that have found lexical diversity differences between L1 and L2 writers (González, 2017; Treffers-Daller, 2013; Yu, 2020).

These results might reflect the task or prompt we used. The prompt did not draw on specific knowledge but is aimed at drawing on writers' general understanding. It may not have provided opportunity for writers to demonstrate higher levels of lexical diversity, thus causing the results to differ from studies that have found differences in lexical diversity between human writing across different levels of education (González López & López-López, 2015; Martins, 2016). Another consideration is time. Previous research has indicated that writing duration may be positively related to lexical diversity (Woods et al., 2023). While there was no time limit for the written responses, participants were compensated via flat rate and may have worked to finish the survey; as mentioned previously, however, the platform automatically calculates average response time and applies retroactive bonuses when the average increases beyond a minimum threshold of $8 USD per hour to mitigate these effects. As some previous studies have failed to find group differences in lexical diversity between L1 and L2 writers (e.g., Kendro & Jarvis, 2023; Kyle et al., 2021) but not education levels, our results may reflect ceiling effects related to task or prompt.

***Can LLMs produce "human-like" lexical diversity?***

Our final question asked whether any of the ChatGPT models produced texts with human-like lexical diversity values. We found all of the ChatGPT-generated texts had different lexical diversity patterns than the human-written texts, except with respect to disparity. One-way ANOVAs revealed that disparity for ChatGPT-3.5 and ChatGPT-4.5 did not differ significantly from any of the human participant groups, and ChatGPT-o4 mini's disparity patterned similarly to one participant group (L1 English writers with bachelor's degrees). We therefore conclude that, in responding to this prompt, ChatGPT models did not produce "human-like" lexical diversity even when compared to multiple groups of human writers with different L1/L2 status and levels of education. Of note, the newer ChatGPT versions (ChatGPT-4.5 and ChatGPT-o4 mini) produced lexical diversity patterns that are even more different from the human-written texts than earlier versions (GPT-3.5 and ChatGPT-4.0). These results were confirmed by our second SVM analysis. Simply put, the human subjects produced wider ranges for most lexical diversity metrics (i.e., variety-repetition, evenness, disparity, dispersion) than the GPT models, which emphasizes a diversity present in human writing across education levels and language status that the ChatGPT-generated texts fail to replicate in responding to this prompt.

It is important to consider multiple reasons why the ChatGPT models differed from humans in this task. It is clear that for the given prompt ChatGPT models did not produce human-like LD, but with further prompt engineering or for different genres the texts produced by AI might pattern more closely. Future research should therefore employ a multi-prompt framework to examine whether our results from a single TOEFL prompt reflect ceiling effects versus true systematic differences in lexical diversity produced by human writers and ChatGPT models. However, given the accessibility of the tool and how typical users engage with it, the essays provided by ChatGPT are more likely representative of common use cases. This may be true for both the overall differences as well as the differences between models.

**Conclusion**

The results of our study show that ChatGPT-generated writing patterns differently from human writers when responding to the TOEFL essay prompt, regardless of the lexical diversity measures employed. This held true regardless of the ChatGPT model and across human writers' education levels and L1/L2 status. We found that ChatGPT produced higher lexical diversity than humans, which differs from the findings of Reviriego et al. (2024) and Jiang and Hyland (2025), but is in line with studies that have examined L2 English writers (Nasseri & Thompson, 2021).

All stakeholders must grapple with the growing role of GenAI-generated language (Chapelle, 2025). As these models are developed, researchers must continue to carry out systematic analyses and share their findings regarding how GenAI writing differs from human-generated text, as well as to explore the ways that these models might assist writers. Even though prior research indicates that human judges are generally not able to distinguish between GenAI-and human-produced texts, the two types of writing evaluated in this study were distinguishable based on the analysis performed.

An important point for consideration in the language learning classroom is whether ChatGPT-generated texts are an appropriate model for students in L2 writing classes, given that the ChatGPT texts may pattern dramatically differently than human writing. This is something to consider beyond concerns of educational equity, plagiarism, and the like. Applied linguists have already raised concerns about the use of AI-written texts, since if such writing patterns differently from what humans would write, students could be exposed to language that is filtered through a machine, or ‘imitating’ language (Warschauer et al., 2023), which may have far-reaching implications for the evolution of language itself (Martínez et al., 2024), as some have found an impact on academic writing already (Bao et al., 2025). Alternatively, the texts produced by ChatGPT might be found to represent superior writing, in which case its use might present models to aspire to imitate, or at least to help learners improve.

Concerning future directions, literature has reported that writers who utilize LLM tools in the writing process have higher levels of lexical diversity (Kim & Chon, 2025; Lin et al., 2025). This also presents possible support for the impact of task type, as some have found it to affect outcomes for GenAI-writing (Fernández-Mira et al., 2021). Future research may consider including GenAI-assisted writing by both L1 and L2 writers, as doing so may show an equalizing effect for writers from both backgrounds (Lin et al., 2025), in combination with a more complex writing task. Our SVM results suggest another avenue for future research. Although our SVM analysis was intended to address our research questions by expanding our statistical analysis beyond the comparison of group means, the model’s strong ability to distinguish between texts written by humans vs. ChatGPT suggests that a model such as this could be a valuable addition to existing AI detection tools.

Crucially, LLMs are continually refined and developed. Over the course of our data collection, OpenAI released ChatGPT-4.5 for research preview along with ChatGPT-4.0 and ChatGPT-4.1. Output from the models we analyzed (ChatGPT-3.5, ChatGPT-4.0, ChatGPT-o4 mini, and ChatGPT-4.5) showed some differences, often minor, in lexical diversity; however, they still did not appear human-like in these metrics. They represent nearly 2.5 years of LLM refinement, indicating that, for now, producing human-like levels of lexical diversity may still elude the most advanced models. As GenAI is increasingly introduced into facets of human experience, we must remember that, for now, their output remains qualitatively and quantitatively different from human writing.

Appendix

Table A1. Descriptive statistics of the LLM- vs. human-generated texts for all lexical diversity measures

| LD measure | Writer type | Mean | CI mean | SD | Minimum | Maximum |
|---|---|---|---|---|---|---|
| Volume | LLM (*n* = 120) | 465.53 | [449.53, 481.53] | 88.51 | 284.00 | 675.00 |
| | human (*n* = 240) | 273.01 | [268.78, 277.24] | 33.26 | 247.00 | 490.00 |
| Abundance | LLM | 249.32 | [240.40, 258.24] | 49.35 | 178.00 | 393.00 |
| | human | 130.97 | [128.42, 133.52] | 20.03 | 92.00 | 215.00 |
| MATTR | LLM | 42.99 | [42.67, 43.30] | 1.75 | 40.23 | 46.16 |
| | human | 38.49 | [38.25, 38.73] | 1.89 | 33.17 | 43.53 |
| Evenness | LLM | .98 | [.98, .98] | .01 | .97 | .99 |
| | human | .97 | [.97, .97] | .01 | .96 | .99 |
| Disparity | LLM | 1.04 | [1.04, 1.04] | .01 | 1.03 | 1.06 |
| | human | 1.03 | [1.03, 1.03] | .01 | 1.01 | 1.06 |
| Dispersion | LLM | 6.87 | [6.52, 7.23] | 1.98 | 2.16 | 10.78 |
| | human | 16.39 | [15.86, 16.92] | 4.14 | 5.43 | 28.96 |

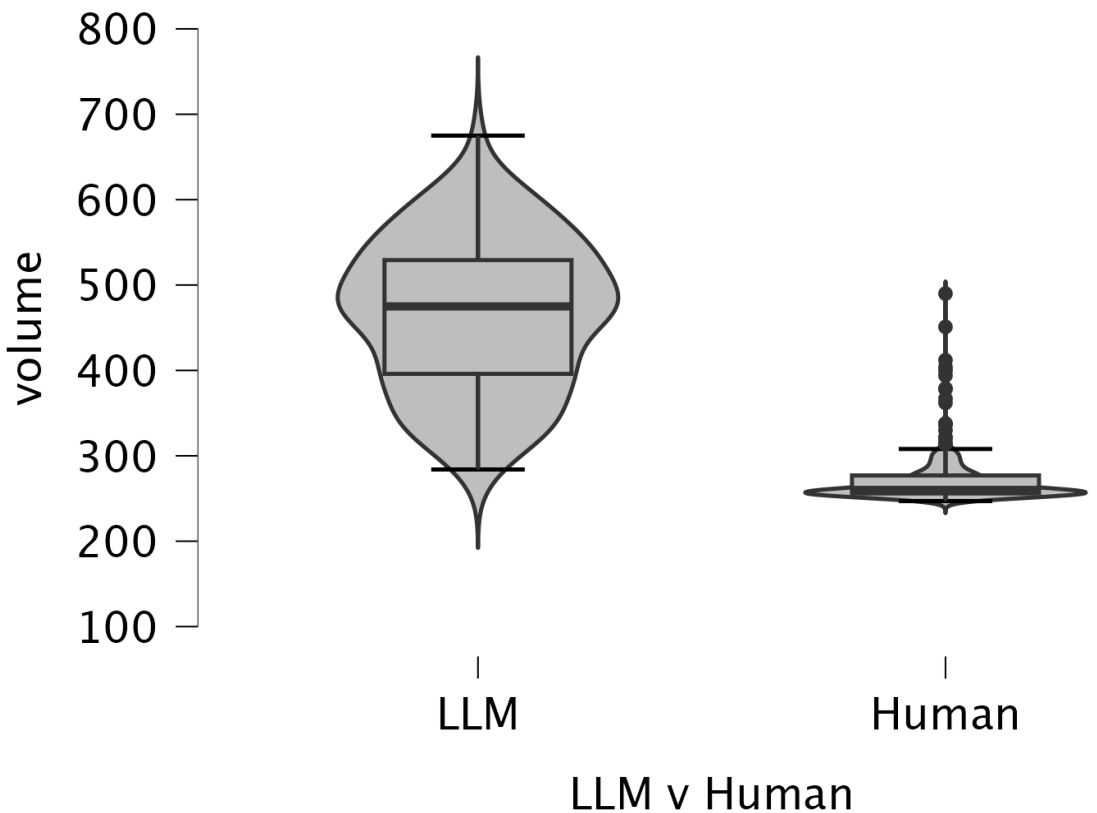

Figure A1. LLM vs. human distributions for volume (i.e., number of tokens)

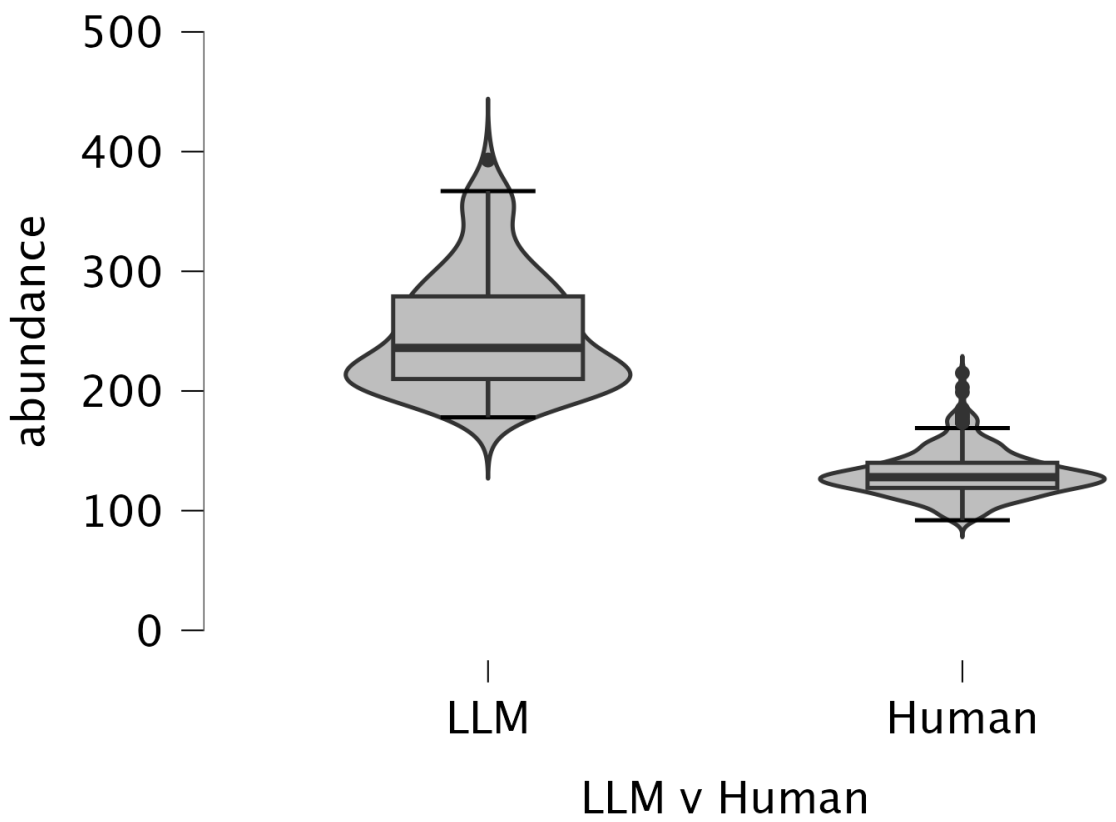

Figure A2. LLM vs. human distributions for abundance (i.e., number of types)

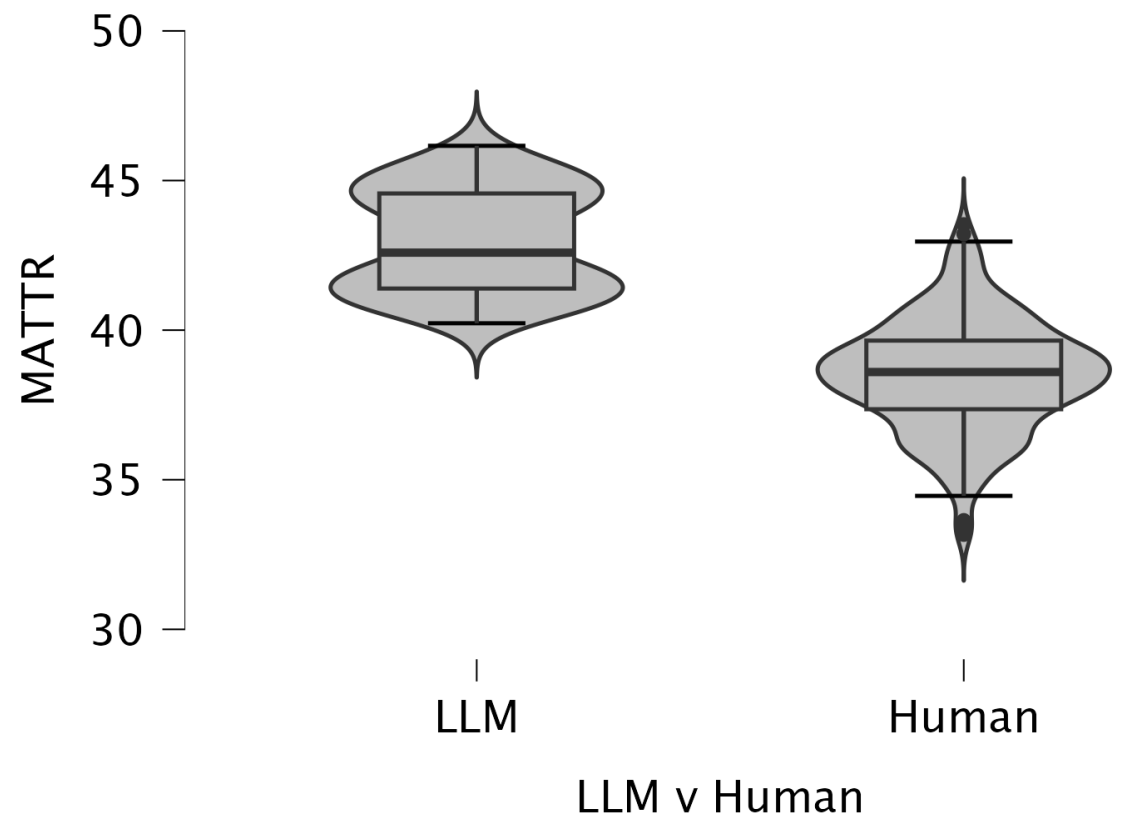

Figure A3. LLM vs. human distributions for MATTR (i.e., variety–repetition)

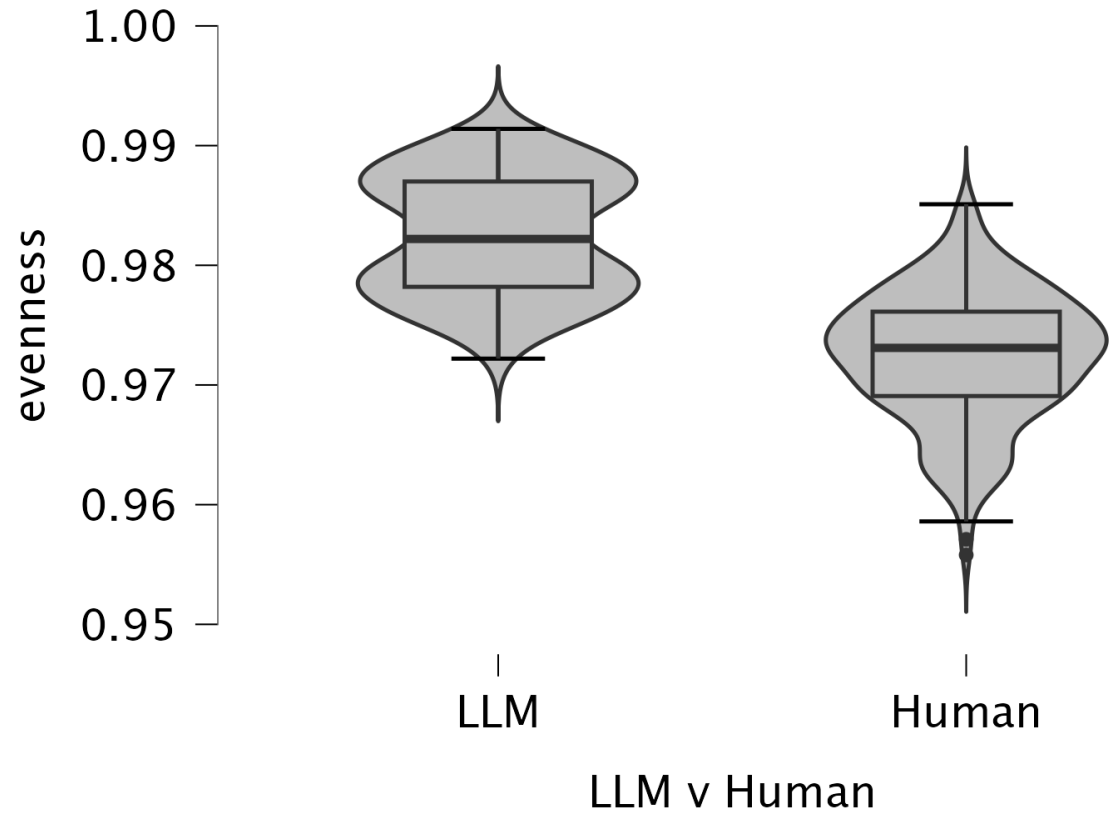

Figure A4. LLM vs. human distributions for evenness (i.e., degree to which different types are equally represented in the text)

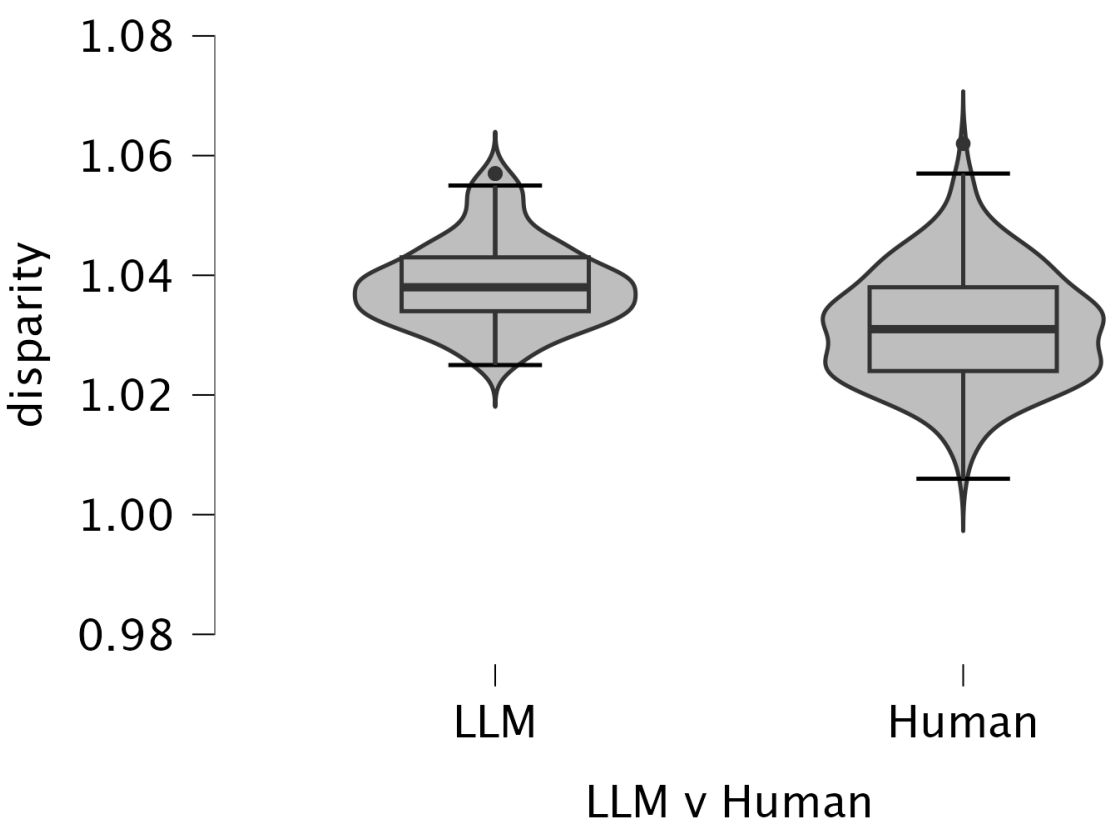

Figure A5. LLM vs. human distributions for disparity (i.e., proportion of related words in the text)

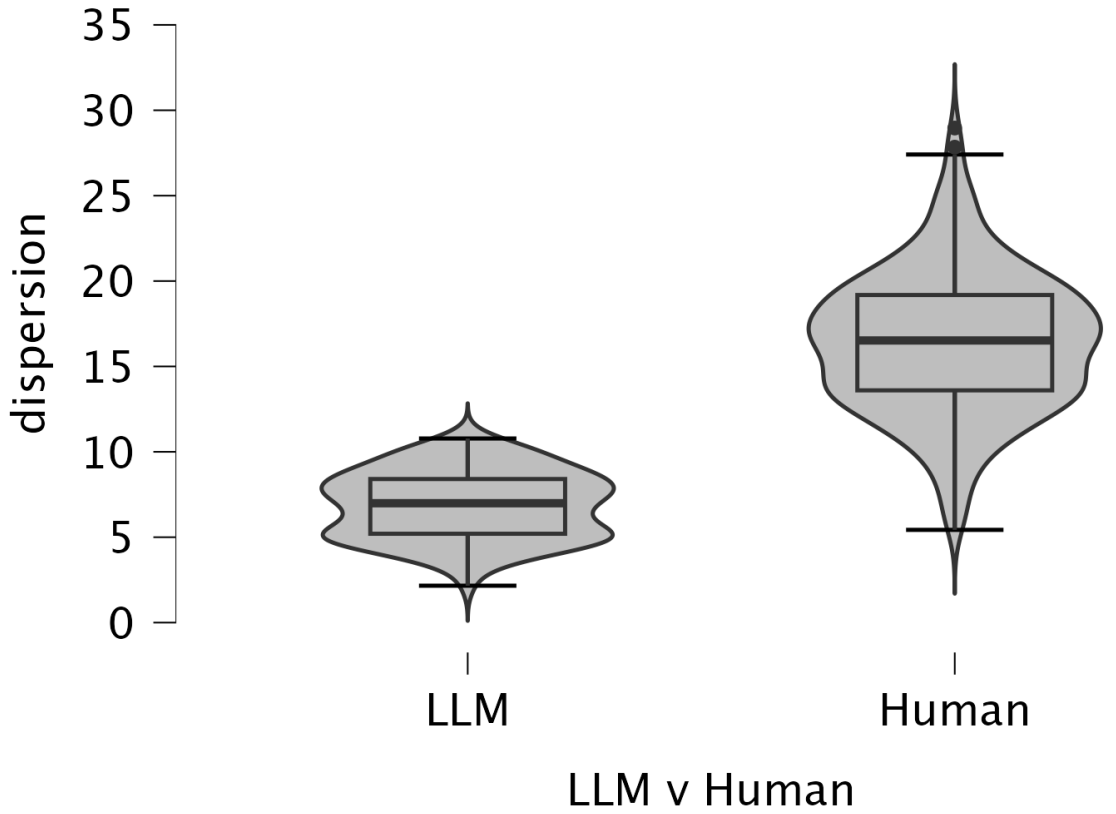

Figure A6. LLM vs. human distributions for dispersion (i.e., repetitions in close proximity)